\documentclass{article}

\usepackage{algpseudocode}

\usepackage{graphicx}
\usepackage{wrapfig}   
\usepackage{lipsum} 
\usepackage{amsfonts} 
\usepackage{booktabs}

\usepackage{titlesec} 

\usepackage[preprint]{corl_2026} 

\usepackage{booktabs}
 \usepackage{amssymb}

\usepackage{colortbl}
\usepackage{xcolor}

\usepackage{caption}
\usepackage{subcaption}

\usepackage{algorithm}

\definecolor{tableheader}{RGB}{100, 200, 230}  

\usepackage{amsmath}
\usepackage{xcolor}
\usepackage{soul}
\usepackage{enumitem}

\makeatletter
\renewcommand{\section}{\@startsection{section}{1}{\z@}%
  {-0.5em}{0.3em}{\large\bf\raggedright}}
\renewcommand{\subsection}{\@startsection{subsection}{2}{\z@}%
  {-0.2em}{0.15em}{\normalsize\bf\raggedright}}
\renewcommand{\subsubsection}{\@startsection{subsubsection}{3}{\z@}%
  {-0.2em}{0.15em}{\normalsize\bf\raggedright}}
\renewcommand{\paragraph}{\@startsection{paragraph}{4}{\z@}%
  {-0.2em}{-0em}{\normalsize\bf\raggedright}}
\makeatother

\title{Scale Up Strategically: Learning Compositional Generalization via Bias-Aware Evaluation and Data Collection for Robotic Manipulation}

\author{
  Yu Qi$^{1}$ , Zhang Ye$^{1}$ , Xinyi Xu$^{1}$ , Yuxuan Lu$^{1}$ , Amitoj Sandhu$^{2}$ , \\
  \textbf{Boce Hu$^{1}$} , \textbf{Haojie Huang$^{1}$} , \textbf{Jonathan Tremblay$^{2\dagger}$} , \textbf{Lawson L.S. Wong}$^{1\dagger}$ \\[4pt]
  $^{1}$Northeastern University \qquad $^{2}$NVIDIA \\[2pt]
  $\dagger$Equal Advising
}

\begin{document}
\maketitle

\vspace{-6mm}



\begin{abstract} Compositional generalization is essential for robot to follow diverse instructions. However, pretrained policies are known to take shortcuts, deferring to salient cues rather than grounding language. 
We introduce a diagnostic framework that localizes this failure to individual \textit{instruction factors}, \textit{e.g.,} reusable semantic components such as color, verb, object, size, and spatial attribute. 
Our framework formalizes instruction factor bias, the tendency of fine-tuned policies to over-rely on dominant factors as shortcuts, and quantifies it through two metrics: Factor Dominance Rate (FDR), capturing pairwise bias between factors, and Factor Dominance Hierarchy (FDH), aggregating these into a global ranking. Evaluation on six foundation policies reveals broadly consistent ordering, \textit{i.e.}, color $\geq$ object $\geq$ spatial $\geq$ verb $\geq$ size, with color dominant, and verb and size most under-grounded. 
We further show the diagnosis is actionable: a bias-aware data collection strategy that reallocates a fixed budget toward under-grounded factors outperforms baselines in simulation and on a real robot using half the demonstrations, thereby enabling more sample-efficient and generalizable policy learning.


\end{abstract}

\vspace{-2mm}


\begin{figure}[h]
    \centering
    \includegraphics[width=\linewidth, ]{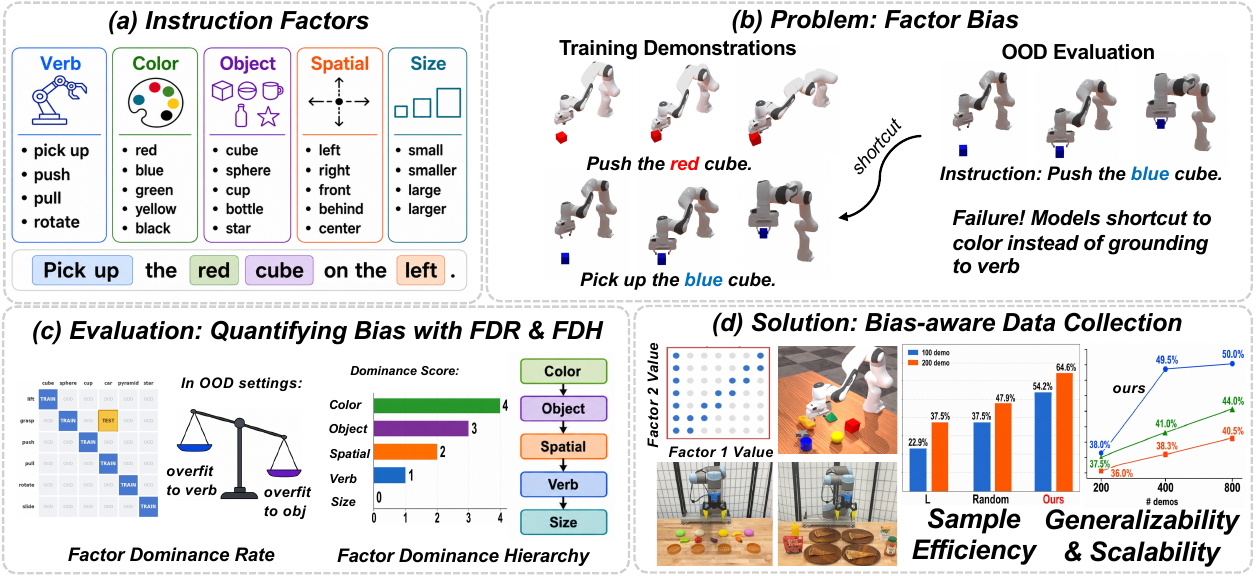}
    \caption{(a) We decompose language instructions into factors. 
    (b) We identify the problem of \textbf{factor bias}: during finetuning, pretrained models tend to overfit to the training data containing certain factors compared with others. 
    Such as models overfit to ``blue'' (color) for picking up the cube instead of pushing it (verb). 
    (c) We quantify this bias with Factor Dominance Rate (FDR) for pairwise factor evaluation and Factor Dominance Hierarchy (FDH) for global ranking. 
    (d) We introduce diagnosis-guided bias-aware data collection for sample-efficient and generalizable policy learning. 
    }
    \label{fig:overview}
\end{figure}

\section{Introduction}

Scaling data is the dominant recipe for generalization in modern robot learning, yet more data does not reliably teach a policy to follow language. 
A general robotics policy trained on large, diverse demonstrations often succeeds on a command not because it parsed the words, 
but because it latched onto whatever cue most reliably predicted the right action. 
For example, in Figure~\ref{fig:overview} (b), the policy ignores the action verb and uses a shortcut with respect to the object-type-action relationship. 
Such behavior is well documented across machine learning~\cite{geirhos2020shortcut, lake2018generalization} and increasingly seen in robot policies. 


Such shortcuts have been studied in robot foundation policies, which are observed to defer to visually salient cues rather than grounding the instruction~\citep{agrawal2018don,fang2026vision}. 
However, existing analyses remain coarse, and they largely report aggregate success rates under varied conditions:
dataset diversity~\citep{xing2025shortcut,lin2025data}, language perturbation~\citep{fei2025libero,liu2023libero}, and visual robustness~\citep{choi2026vla,liu2025roboview}. 
This reveals whether a policy fails but not why. 
As a result, it remains unclear which parts of an instruction a policy actually relies on, how strongly, and how to repair the failure.





In this work, we systematically study these shortcut behaviors at the level of \textbf{instruction factors}, which we define as reusable and independent semantic components of an instruction, such as descriptive color, action verbs, object types, \textit{etc}. 
Fig.~\ref{fig:overview} shows that policies generalize unevenly across factors in out-of-distribution settings. 
We refer to such failure as \textbf{instruction factor bias}: fine-tuned policies tend to over-rely on dominant factors as shortcuts while under-grounding others. 
To evaluate this bias, we construct a structured benchmark over five common dimensions for everyday robotics tabletop instructions: 
color, verb, object, size, and spatial attribute. 
To provide in-depth analysis, we introduce two diagnostic metrics: \textbf{Factor Dominance Rate (FDR)}, which measures pairwise factor bias, and \textbf{Factor Dominance Hierarchy (FDH)}, which aggregates pairwise FDR scores into a global ranking.
Finally, across six foundation policies, the factor bias is strikingly consistent: color is the most dominant factor, verb and size the most under-grounded, and the induced hierarchy is: color $\geq$ object $\geq$ spatial $\geq$ verb $\geq$ size holds broadly across multiple architectures.

One of the practical values of this diagnosis lies in turning shortcut diagnosis into a bias-aware data collection principle. 
Since the instruction space grows combinatorially with the number of factors, exhaustive coverage is often infeasible, especially in real-robot settings. 
We propose a model-agnostic strategy that reallocates a limited demonstration budget toward under-grounded factors, \textit{i.e.}, those ranked lower by FDH. 
Our strategy outperforms baselines in most simulation and real-robot settings; on the real robot, it achieves stronger performance with half the demonstrations. 
Together, these results show that beyond increasing data quantity and diversity, shaping the data distribution to mitigate factor bias can more effectively improve compositional generalization.

In summary, our contributions are threefold: 
\vspace{-1mm}
\begin{itemize}[leftmargin=*, topsep=1pt, itemsep=3pt, parsep=0pt]
  \item We identify \textbf{factor bias} in language instructions as a 
        systematic and previously unmeasured failure mode that bottlenecks 
        compositional language generalization in pretrained robot policies.
  \item We introduce the first quantitative framework for factor bias 
        evaluation, and propose \textbf{FDR} and \textbf{FDH}, revealing 
        consistent findings shared across pretrained policy backbones.
  \item We propose a simple yet effective \textbf{bias-aware data collection 
        strategy} that outperforms baselines under matched demonstration 
        budgets, validated through simulation and real-robot experiments 
        across multiple models.
\end{itemize}
\vspace{-1mm}

\section{Related Works}

\paragraph{Compositional Generalization in Robotic Manipulation. } We study compositional generalization~\citep{lake2018generalization,hupkes2020compositionality,keysers2019measuring,lin2023survey,li2024context} of language instructions--the ability to recombine learned primitives for novel instruction combinations. Prior works pursue this through explicit structural priors (modular architectures~\citep{devin2017learning}, symbolic planners~\citep{xu2018neural,kuo2020deep}, programmatic policies~\citep{qiu2022programmatic}) or implicitly via large-scale multitask learning in models such as RT-1/2~\citep{brohan2022rt,zitkovich2023rt} and Openpi~\citep{black2024pi_0,intelligence2025pi} trained on DROID~\citep{khazatsky2024droid} and Open X-Embodiment~\citep{o2024open}. In this work, we for the first time identify and evaluate \textbf{factor bias} in language instructions as a systematic and previously-unmeasured failure mode that bottlenecks compositional generalization in pretrained robot policies.

\paragraph{Evaluation of Shortcut Behaviors in Robot Foundation Models. } Prior works observe that pretrained robot policies exhibit shortcut behaviors, deferring to visually salient cues rather than grounding in instructions~\citep{agrawal2018don,cadene2019rubi,fang2026vision,o2024open,khazatsky2024droid}. Our evaluation differs in two respects. First, prior studies measure success rate under dataset diversity~\citep{xing2025shortcut,lin2025data,shi2026diversity}, background texture~\citep{wu2025policy}, language perturbation~\citep{fei2025libero,liu2023libero}, and visual robustness~\citep{choi2026vla,liu2025roboview}, yet fine-grained shortcut analysis along instruction factors remains under-explored. Second, they typically report task success rate~\citep{fei2025libero,liu2023libero,chen2025robotwin,nasiriany2024robocasa,mees2022calvin,yang2026robolab} in simulation rather than diagnosing failure behaviors—revealing whether a policy fails but not why, making actionable improvements hard to derive. We fill this gap in the literature with a systematic framework for evaluating factor-bias failure behaviors, the FDR and FDH evaluation metrics, and a bias-aware data collection strategy for compositional generalization.

\paragraph{Data Collection Strategy in Robotic Manipulation. }

\begin{wrapfigure}[9]{r}{0.25\linewidth}  
\vspace{-18pt}
    \centering
    \includegraphics[width=\linewidth]{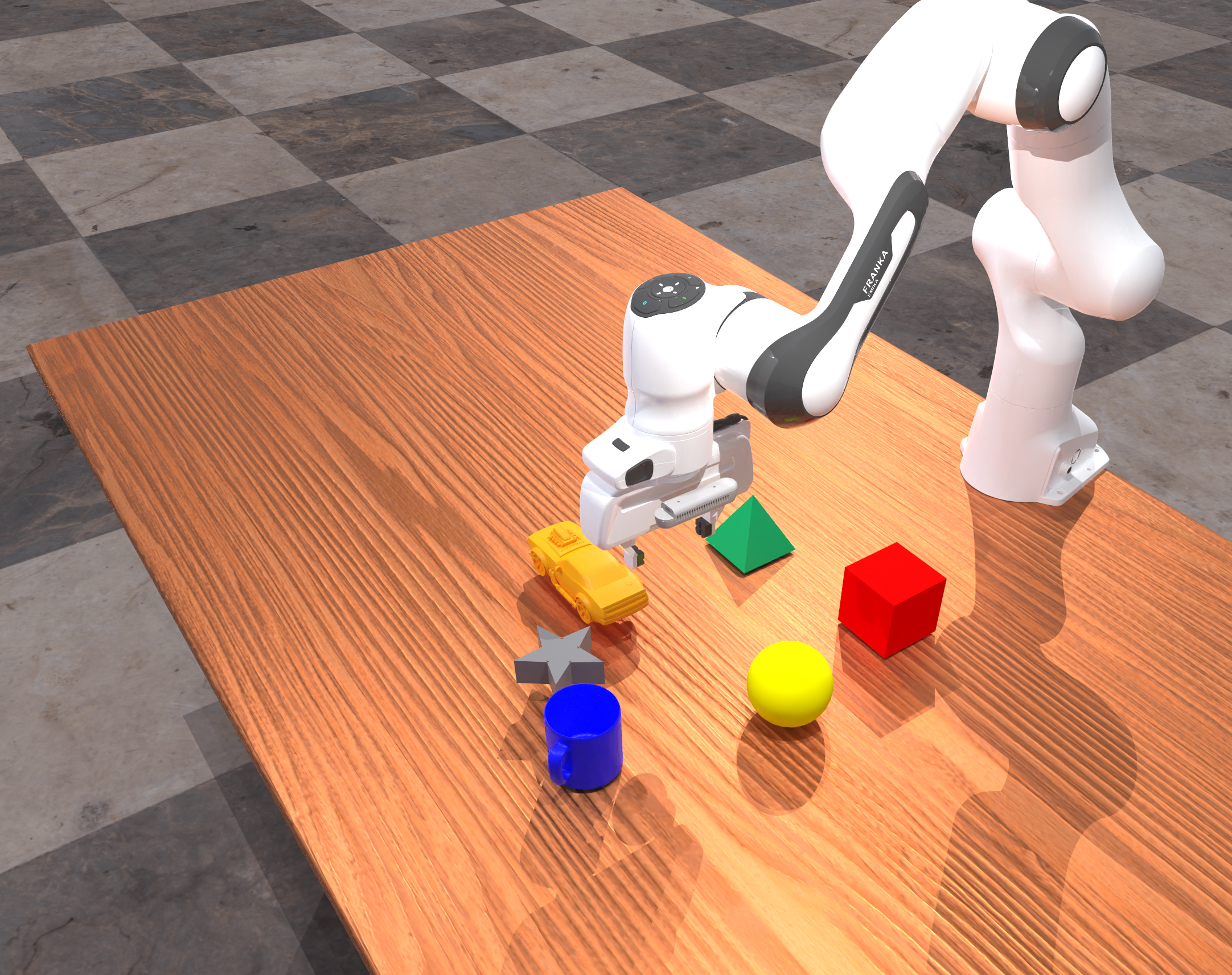}
    \caption{Simulation setup in Maniskill.}
    \label{fig:setup}
\vspace{-12pt}
\end{wrapfigure}

Collecting robot demonstrations at scale remains a major bottleneck in imitation learning~\citep{mandlekar2018roboturk,kelly2019hg}. A large body of work reduces this cost by improving data acquisition pipelines, including scalable teleoperation, motion capture, and sim-and-real co-training, active learning~\citep{wang2024dexcap,lepert2025phantom,maddukuri2025sim,saxena2025matters, salganicoff1996active}. Complementary to these collection efficiency efforts, other studies argue that task coverage—not dataset size—is often the real bottleneck to generalization~\citep{belkhale2023data,lin2025data,xing2025shortcut}. Building on compositional generalization over environmental factors~\citep{gao2024efficient} and motivated by our factor bias diagnosis, we propose a \textbf{bias-aware data collection strategy} that allocates demonstrations toward under-represented factors to improve compositional generalization.





\section{Problem Formulation}
\label{problem_formulation}
As shown in Fig.~\ref{fig:setup}, our experiments are conducted in a multi-task tabletop manipulation setup in ManiSkill~\citep{mu2021maniskill}, where each natural-language instruction is decomposed into five factors following the template ``\textit{[Verb] the [Size] [Color] [Object] on the [Spatial Attribute] of the table}'' (e.g., \textit{``Grasp the smallest red cube on the left of the table''}). 

\textbf{The Factor Vocabulary} consists of five common factors of everyday instructions in robotic manipulation:  \textit{\textbf{Verb}} (\textit{grasp, lift, push, pull, rotate, slide}), \textit{\textbf{Color}} (\textit{red, yellow, blue, orange, green, black}), \textit{\textbf{Object}} (\textit{cube, sphere, cup, car, pyramid, star}), \textit{\textbf{Size}} (\textit{small, large, smaller, larger, smallest, largest}), and \textit{\textbf{Spatial Attribute}} (\textit{left, right, middle, front, behind}). Their Cartesian product defines an instruction space of $6{\times}6{\times}6{\times}6{\times}5=6{,}480$ unique instructions; each scene adds up to three random distractors with random spatial configurations. Policies are trained and evaluated on this instruction space or subspace. More detailed experiment settings provided in the supplementary materials.

\section{Evaluation of Factor Bias}
\label{section_evaluation_factor_bias}

\begin{figure}[t]
    \centering
\includegraphics[width=\linewidth, height=5cm, keepaspectratio]{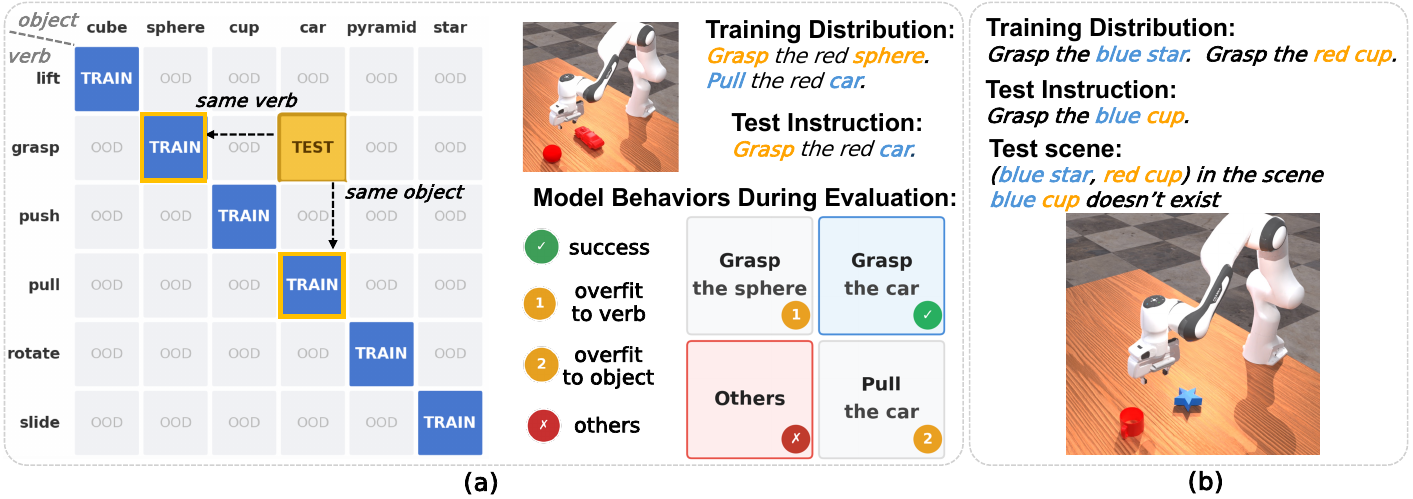}
\caption{\textbf{Factor correlation experiment setups.} \textbf{(a) The (verb, object) 
setting.} 
During inference, only two objects are in the scene with the same color: the target object(``car'') and the verb-paired object during training(``sphere''). We compute FDR by checking whether it overfits
to the training-paired verb or to the training-paired object. \textbf{(b) The (color, object) setting} follows similar training and inference setup; 
at inference, the instructed object does not exist in the scene.
}

\vspace{-4mm}    
\label{fig:factor_correlation_experiment_setups_visualization}
\end{figure}

This section introduces our factor-level diagnostic framework. 
We first define Factor Dominance Rate (FDR) to measure pairwise factor dominance, and then aggregate pairwise scores into Factor Dominance Hierarchy (FDH) to reveal global bias patterns across models.


\subsection{Factor Dominance Rate (FDR) for Pairwise Factor Bias Evaluation}

\label{factor_dominance_rate_for_pairwise_factor}

\paragraph{Setup. } To probe the bias within a factor pair $(f_1, f_2)$, we finetune a policy on a
training distribution that deliberately correlates the two factors (In Fig.~\ref{fig:factor_correlation_experiment_setups_visualization}~(a), each value of $f_1$ is paired
with one value of $f_2$ (blue cells)) while
randomizing all other factors and spatial layouts. We then evaluate on the unseen off-diagonal
combinations (gray cells) and observe which factor drives the resulting overfitting.


\paragraph{Construction of the inference scene. } 

The setup of the scene is determined by the factor type. 
We distinguish an \textbf{action-centric} factor (the \textit{verb}), which determines how the
  policy interacts with an object, from \textbf{object-centric} factors (\textit{color}, \textit{object}, \textit{size},
  \textit{spatial attribute}), determining which object is selected and where. This gives two cases, shown in Fig.~\ref{fig:factor_correlation_experiment_setups_visualization}~(a) and (b):
  \begin{itemize}[leftmargin=*, topsep=2pt, itemsep=2pt, parsep=0pt]
\item \textbf{Verb vs.\ an object-centric factor} (Fig.~\ref{fig:factor_correlation_experiment_setups_visualization}~(a)).
The scene presents two objects sharing the same color: the instructed target (e.g., \textit{car''}) and the object the instructed verb was paired with in training (e.g., \textit{sphere''} for \textit{grasp''}). Overfitting to the verb manifests as acting on the verb-paired object (grasping the sphere), whereas overfitting to the object manifests as applying that object's training verb (\textit{pull''}) to the instructed target (\textit{``car''}).
    \item \textbf{Two object-centric factors} (Fig.~\ref{fig:factor_correlation_experiment_setups_visualization}~(b)). We issue
  a deliberately impossible instruction (e.g., \textit{``grasp the blue cup''}) into a scene containing only the two
  training-paired objects (a \textit{blue star} and a \textit{red cup}). 
As no single object satisfies both attributes, the one the policy acts on indicates whether it is driven by color or by object.
   Reaching for the \textit{blue star} means it followed
   the color while ignoring the object;
  reaching for the \textit{red cup} means it followed the object while ignoring the color.
\end{itemize}

\paragraph{Scoring. } 
%
Rule-based success captures only task completion and offers no reliable way to attribute overfitting failures to a particular factor.
We therefore use Gemini-2.5-Flash~\citep{comanici2025gemini} to classify each
agent-view rollout as success, overfitting to $f_1$, or overfitting to $f_2$,
choosing it as the judge model to balance evaluation efficiency and quality
given the large number of rollout videos. In the supplementary material, we
show that its judgments are highly consistent with human annotations. Finaly we calculte the scoring with
$N_{f_1}, N_{f_2}$ the counts of the two overfitting modes, thus FDR is calculated as:
\begin{equation}
    \text{FDR}(f_1, f_2) = \frac{N_{f_1} - N_{f_2}}{N_{f_1} + N_{f_2} + \epsilon} \in (-1, 1),
    \qquad \epsilon \rightarrow 0 .
\end{equation}

The sign of FDR indicates the direction of factor dominance: positive values indicate dominance by $f_1$, while negative values indicate dominance by $f_2$. Its magnitude reflects the strength of the bias. Values near $0$ suggest that the two factors are grounded in a balanced manner.

\paragraph{FDR Experiment Results. } In Tab.~\ref{fdr_result_table},  we report the experiment results on six robotic policy models. For each pairwise FDR score, 300 demonstrations are used for training, and 400 rollouts are uniformly sampled across OOD cells for the model's evaluation. These scores are further used to calculate FDH in the next section.

\begin{table}[t]
\centering
\small
\setlength{\tabcolsep}{1.25pt}
\caption{\textbf{Factor Dominance Rate (FDR, \%)}, for pairwise factor dominance evaluation
across VLA and video action model backbones. Each FDR is calculated by 400 rollouts uniformly sampled in the OOD distribution. A positive value indicates the first factor 
dominates; a negative value indicates the second factor dominates. 
Values closer to zero indicate more balanced grounding.}
\label{tab_1:fdr_pairwise}
\begin{tabular}{lcccccc}
\toprule
\textbf{Factor Pair} & $\boldsymbol{\pi_{0.5}}$~\cite{intelligence2025pi} & $\boldsymbol{\pi_{0}}$~\cite{black2024pi_0}  & \textbf{OpenVLA-oft}~\cite{kim2024openvla} & \textbf{GR00T-N1.7}~\cite{bjorck2025gr00t} & \textbf{XVLA}~\cite{zheng2025x} & \textbf{Genie-Envisioner}~\cite{liao2025genie} \\
\midrule
(verb, color)     & -4.0 & -6.0 & -6.2 & -6.8 & -1.0 & -22.0 \\
(verb, object)    & -9.2 & -16.7 & 16.0 & -25.3 & -19.0 & -25.0 \\
(verb, size)      & 26.3 & 41.3 & 4.5 & 25.1 & 38.0 & 32.6 \\
(verb, spatial)   & -7.4 & -2.1 & 4.5 & -12.2 & -5.0 & -14.2 \\
(color, object)   & 57.3 & 42.1 & 17.5 & 22.8 & 31.0 & 22.2 \\
(color, size)     & 54.0 & 52.4 & 33.6 & 52.4 & 56.0 & 30.9 \\
(color, spatial)  & 38.6 & 32.6 & 39.5 & 8.4 & 19.0 & 36.5 \\
(object, size)    & 22.0 & 31.2 & 1.6 & 52.4 & 29.0 & 11.4 \\
(object, spatial) & -3.8 & -2.9 & 17.0 & 9.5 & 9.0 & 20.9 \\
(size, spatial)   & -7.9 & -13.0 & 19.8 & -11.0 & -32.0 & -23.7 \\
\bottomrule
\end{tabular}
\label{fdr_result_table}
\vspace{-6.5mm}
\end{table}

\subsection{Factor Dominance Hierarchy (FDH) and Findings} 
\label{fdh_and_findings}


We aggregate the pairwise FDR scores into a global ranking, the Factor Dominance Hierarchy (FDH), via Copeland ranking: each factor $f_i$ receives a Copeland score: $C(f_i) = \sum_{j \neq i} \mathbb{1}\!\left[\mathrm{FDR}(f_i, f_j) > \tau\right]$ where $\tau$ is a tie threshold (we use $\tau = 5\%$). A factor scores $+1$ for each pairwise win, and $0$ for each loss or each tie.
Factors are ranked in descending order of $C(f_i)$. 
We report the FDH of all models in Tab.~\ref{tab:fdh}, and observe the following three findings: 

\begin{table}[t]
\caption{\textbf{Factor Dominance Hierarchy (FDH)}, for each model, derived via Copeland ranking over pairwise FDR scores. $>$ indicates strict dominance; $\approx$ indicates comparable performance (tie).}  
\label{tab:fdh}
\centering
\begin{tabular}{lp{0.40\linewidth}}
\toprule
\textbf{Model} & \textbf{Factor Dominance Hierarchy (FDH)} \\
\midrule
$\boldsymbol{\pi_{0.5}}$~\cite{intelligence2025pi} & color $>$ object $\approx$ spatial $>$ verb $>$ size \\
$\boldsymbol{\pi_{0}}$ ~\cite{black2024pi_0}     & color $>$ object $>$ spatial $\approx$ verb $>$ size \\
OpenVLA-oft~\cite{kim2024openvla} & color $>$ object $\approx$ verb $\approx$ size $>$ spatial \\
GR00T-N1.7~\cite{bjorck2025gr00t}  & color $>$ object $>$ spatial $>$ verb $>$ size \\
XVLA~\cite{zheng2025x} & color $\approx$ object $>$ spatial $>$ verb $>$ size \\
Genie-Envisioner~\cite{liao2025genie} & color $>$ object $>$ spatial $>$ verb $>$ size \\

\bottomrule
\vspace{-5mm}
\end{tabular}
\label{fdh_table}
\end{table}

\sethlcolor{blue!8}   
\hl{\textbf{\textit{Finding 1: Color is the most dominant instruction factor.}}} Across all evaluated backbones, color consistently emerges as the 
most dominant factor. One possible explanation is that color constitutes a highly salient low-level visual cue: it is reliably captured by vision encoders, and most likely to induce shortcut correlations in large-scale training data.

\sethlcolor{blue!8}
\hl{\textbf{\textit{Finding 2: Verb and size are relatively overlooked.}}} Compared to color, verb and size factors exhibit the weakest dominance in most backbones, except for OpenVLA-oft, which shows stronger verb and size grounding. One possible explanation is that these factors are harder to learn consistently from large-scale demonstrations, especially since size often depends on relative geometric relationships, and existing datasets may provide insufficient diversity in this regard.

\sethlcolor{blue!8}
\hl{\textbf{\textit{Finding 3: Different models share FDH consistency.}}} In Tab.~\ref{fdh_table}, except for OpenVLA-oft~\cite{kim2024openvla}, the factor hierarchy is broadly consistent across architectures: \textbf{color $\geq$ object $\geq$ spatial $\geq$ verb $\geq$ size}. This consistency indicates that the factor bias hierarchy is largely a
shared property of multiple policy backbones, which motivates a model-agnostic data collection strategy, described in Sec.~\ref{section_data_collection}.

\section{Strategic Data Collection with Factor Bias}
\label{section_data_collection}

\begin{figure}[t]
    \centering
\includegraphics[width=\linewidth]{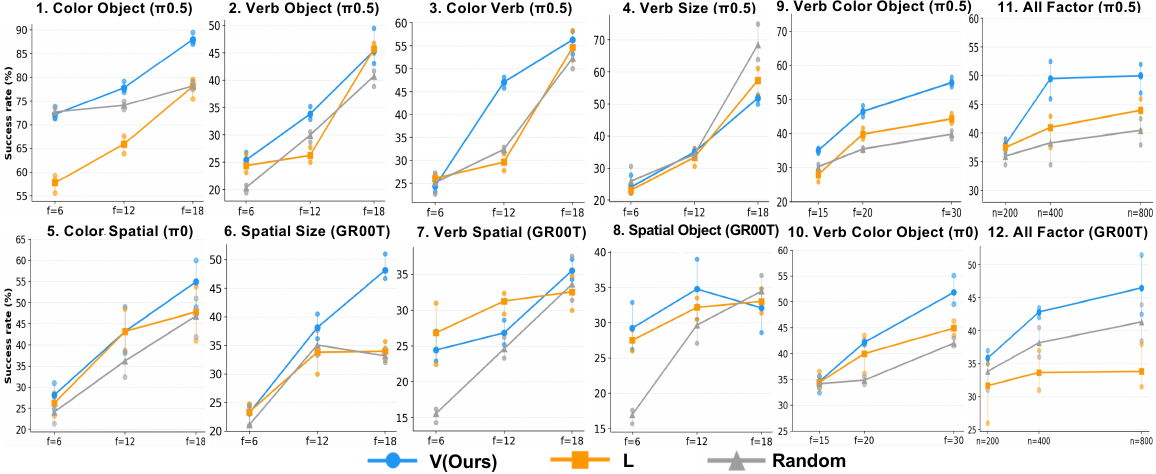}
\vspace{-6pt}
\caption{\textbf{Simulation Experiment Results.} Each panel plots the success rate under factor change $f$
 or demonstration budget $n$. In plot 11, 12, $f=50$, in all other plots, $n=200$. Each checkpoint is evaluated under three random seeds, each seed with 200 rollouts. Dots are per-seed runs, lines represent their mean values. More results and details are provided in the supplementary materials.
}
\vspace{-4mm}    
\label{fig:simulation_experiment_result_new}
\end{figure}

We hypothesis that the consistent factor hierarchy observed in \textit{Finding 3} can directly guide data collection. 
Rather than searching for an optimal sampling strategy globally, our goal is to provide a simple yet effective bias-aware data collection strategy which yields substantial gains in compositional generalization and data efficiency.

\begin{wrapfigure}[10]{r}{0.25\linewidth}
    \vspace{-14pt}
    \centering

    \includegraphics[width=\linewidth]{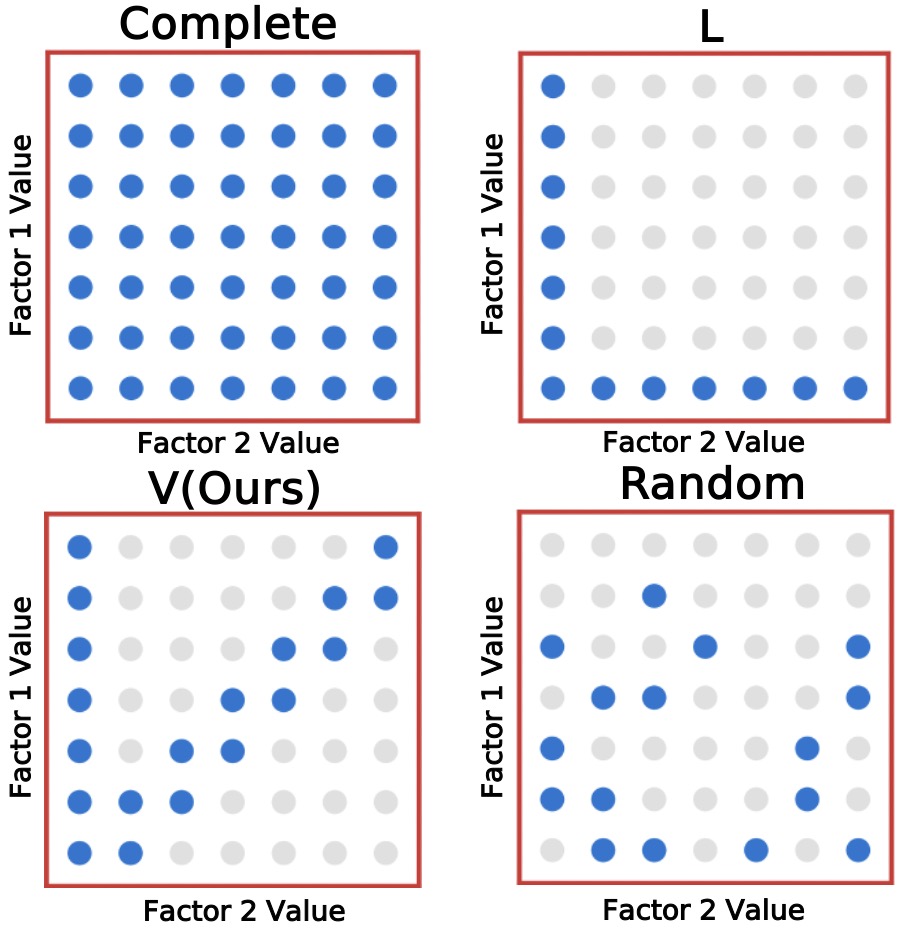}

    \vspace{-2pt}

    \caption{Methods.}
    \label{fig:data_collection_strategy}
    \vspace{-12pt}
\end{wrapfigure}

\paragraph{Method. }    \label{baseline_method} 
Fig.~\ref{fig:data_collection_strategy} visualizes different sampling strategies for 2 factors, 
axes enumerate the possible values of the two factors, 
blue dots mark the samples collected, while gray dots mark the held-out samples. \textbf{\textsf{V} (Ours)} is inspired by \citet{gao2024efficient}  
we collect the leftmost column—fixing $f_1$
while varying $f_2$, where $FDR(f_2, f_1) > \tau$, meaning $f_1$ is the under-grounded factor. In addition, we include the diagonal and subdiagonal axes.
For three or more factors ($N_f\geq3$), a column is added per under-grounded factor, prioritizing the factors ranked lower by FDH. 
\textbf{\textsf{Complete}} strategy exploits all possible combinations and serves as an upper-bound performance reference, which is infeasible, especially in real-robot scenarios. 
\textbf{\textsf{Random}} strategy randomly samples possible factor combinations in the full space. 
\textbf{\textsf{L}} strategy fixes each of the factors and samples uniformly the other factor.  
Details provided in supplementary materials.

\section{Experiment}
\label{section_experiment}




To test whether the consistency revealed in \textit{Finding 3}
translates into an effective data collection strategy, we conducted experiments across multiple models in both simulation in Sec.~\ref{simulation_experiment_section} and real-robot experiments in Sec.~\ref{real_robot_experiment_section}. 
Additionally, following the exact setup as \citet{gao2024efficient}. We keep the sum of total factor change $f$ and demonstration budget $n$ the same across methods for fair comparisons.

\textbf{Selection of Under-grounded Factors.} To formulate the strategy designed in \textsf{V}, we reuse a single global ordering observed in \textit{Finding 3} (color $\geq$
object $\geq$ spatial $\geq$ verb $\geq$ size) rather than re-estimating
bias for every setting. Concretely, when $N_f=2$, we select the single lower-ranked factor; when $N_f=3$, the two lowest-ranked; and when $N_f > 3$, the three lowest-ranked. Factors are always chosen
in ascending order of dominance, i.e., from the weakest upward.

\textbf{Factor Change ($f$).} Factor Change between two factor configurations is calculated by Hamming Distance~\citep{HammingDistance}. The change of one factor value (e.g., swapping a red cube for a blue cube) is counted as 1 factor change. In \citet{gao2024efficient}, $f$ indicates human efforts for environment setup. In our setting, $f$ measures the variety of configurations in the instruction space, and the task-switching cost. When strategies yield different total factor change $f$, we additionally sample random configurations to equalize $f$ across all methods. Details provided in supplementary materials.

\textbf{Demonstration Budget ($n$).} It denotes the total number of demonstrations allowed to be collected.

\subsection{Simulation Experiment}
\label{simulation_experiment_section}

\textbf{Experiment Setup.} We verify the effectiveness of our data collection strategy on $\pi_0$~\citep{black2024pi_0}, $\pi_{0.5}$~\citep{intelligence2025pi}, GR00T-N1.7~\citep{bjorck2025gr00t}. 
We consider pairwise
factor-change experiments, together with Verb-Color-Object and
All-Factor settings. Each task has a separate instruction subspace:
e.g., in Color\,Spatial, color and spatial vary while the
remaining factors are randomized within two values. Distractors and object location randomization are defined in Sec.~\ref{problem_formulation}. Experiment parameters and results are reported in Fig.~\ref{fig:simulation_experiment_result_new}. We provide further details in the supplementary materials.


\textbf{Result and Analysis.} 
As shown in Fig.~\ref{fig:simulation_experiment_result_new}, under matched factor-change and demonstration budgets, \textsf{V} matches or outperforms the \textsf{L} and \textsf{Random} baselines in the large majority of settings. The gains are largest in the face of a strongly dominant factor: on color-dominated pairs, \textsf{V} improves over the best baseline by $+10$ points on Color\,Object and $+7$ points on Color\,Spatial. The advantage is not confined to color---\textsf{V} also lifts spatial--size grounding on GR00T by $+14$ points, and scales better with the demonstration budget in the All\,Factor setting ($n=800$). The main exception is the Verb\,Size setting on $\pi_{0.5}$, where \textsf{V} ($51\%$) trails Random ($68\%$) at $f=18$; we attribute this to size being intrinsically hard to ground from relative geometry (Finding~2), so additional size demonstrations are less reliably beneficial. The gains are especially pronounced when $N_f \geq 3$. In Plots~9--10 (Verb\,Color\,Object), \textsf{V} at $f=20$ already matches the baseline performance reached only at $f=30$, and improves over the best baseline by up to $15$ points at $f=30$. In Plots~11--12 (All\,Factor), \textsf{V} surpasses all baselines using only half the demonstration budget ($n=400$ vs.\ $n=800$), confirming that our proposed strategy is both sample-efficient and scalable.

\subsection{Real-robot Experiment}
\label{real_robot_experiment_section}

\begin{wrapfigure}[11]{l}{0.26\linewidth}  
    \centering
    \includegraphics[width=\linewidth]{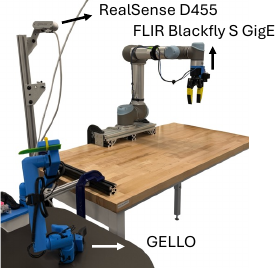}
    \caption{UR5 Setup.}
    \label{fig:real_robot_setup}
\end{wrapfigure}

As shown in Fig.~\ref{fig:real_robot_setup}, we conduct three everyday manipulation tasks: \textit{Bun}, \textit{Pizza}, and \textit{Cup}, on a UR5 robot equipped with a custom-designed gripper, a RealSense D455 as the agent-view camera, and a FLIR in-hand camera. Demonstrations are collected using GELLO~\cite{wu2024gello}. 

\begin{figure}[t]
    \centering
\includegraphics[width=\linewidth]{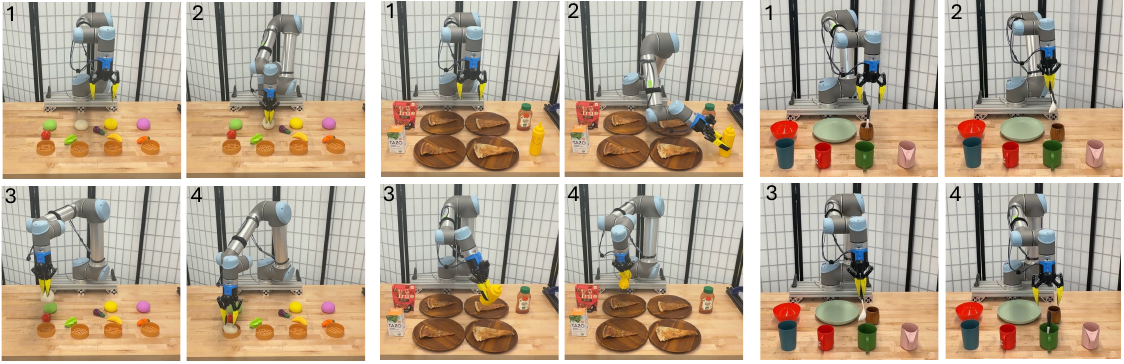}
\vspace{-16pt}
\caption{\textbf{Real-robot Experiment.} We conduct experiments on three manipulation tasks: \textit{Bun}, \textit{Pizza} and \textit{Cup}. Each task contains 16 instruction variants with randomized spatial configurations.}
\label{real_robot_experiment_task}
\vspace{-4mm}
\end{figure}

\paragraph{Task Designs. } As shown in Fig.~\ref{real_robot_experiment_task}, each task is a naturally language-conditioned 
setting with 16 instruction variants formed by jointly varying two 
instruction factors ($f_1$, $f_2$), i.e., a $4\times4$ grid per task. 
\textbf{\textit{Bun}} varies Color\,Spatial: the policy is required to pick 
up the bun of a specified color at a specified location 
(e.g., \textit{``put the green bun into the first steamer basket on the left''}). \textbf{\textit{Pizza}} varies Spatial\,Object: the policy is required to pick up 
a specified seasoning and squeeze it onto the pizza at a fixed location (e.g., \textit{``add mustard sauce to the pizza in the top left corner''}). \textbf{\textit{Cup}} varies Verb\,Color: the policy is required to perform a certain skill on the cup with a specific color (e.g., \textit{``put the stirring spoon into the red cup''}). Details are provided in the supplementary materials.

\paragraph{Experiment Setup. } We evaluate our strategy \textsf{V} against the 
\textsf{Random} and \textsf{L} baselines defined in Sec.~\ref{baseline_method} on GR00T-N1.7~\citep{bjorck2025gr00t}, 
under matched factor change $f=12$ and two demonstration budgets, $n=100$ and $n=200$. Additionally, we include \textsf{Complete} strategy, which 
exhaustively covers the full instruction space, as an upper-bound reference. At inference, each checkpoint is evaluated uniformly over 48 rollouts in total (3 rollouts per instruction). 

\paragraph{Result and Analysis. } We report real-robot experiment results in 
Tab.~\ref{tab:realrobot_experiment_result}. Across all three tasks, \textsf{V} 
consistently outperforms both baselines, improving over \textsf{L} by 
+28.7 points and over \textsf{Random} by +16.7 points on 
average. The advantage holds at both demonstration budgets: on \textit{Bun}, 
\textsf{V} at $n=100$ (54.2\%) already surpasses both \textsf{L} 
(37.5\%) and \textsf{Random} (47.9\%) at $n=200$, showing that bias-aware 
allocation is substantially more sample-efficient: it attains higher success 
with half the demonstrations. As the budget grows from $n=100$ to $n=200$, all 
methods improve while \textsf{V} retains a clear lead. These results transfer 
our simulation findings to a physical platform, confirming that allocating 
demonstrations toward under-grounded factors yield more generalizable, 
data-efficient language-conditioned policies.

\begin{wrapfigure}[9]{l}{0.40\linewidth}  
    \centering
    \includegraphics[width=\linewidth]{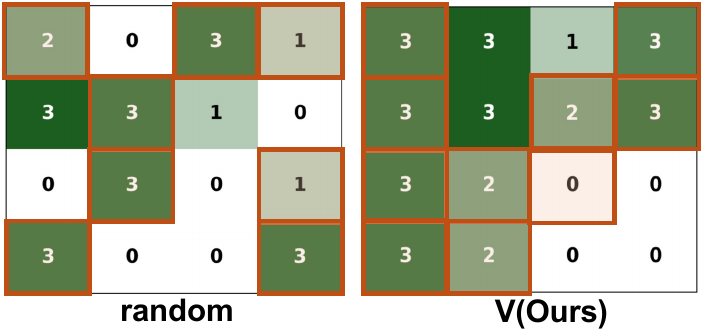}
    \caption{\textit{Bun} results visualization.}
    \label{fig:generalization_visualization_real_robot}
\end{wrapfigure}

\textbf{Partial coverage may yield better performance than full coverage under a small demo budget. } 
In Fig.~\ref{fig:real_world_scaling_behaviors}, our \textsf{V} strategy outperforms \textsf{Complete} under $n=100$ by a large margin. We attribute this to demonstration density: under a fixed small budget, 
\textsf{Complete} spreads demonstrations thinly across the entire combinatorial 
space, leaving each instruction undersampled, whereas \textsf{V} concentrates 
the budget on fewer combinations, providing enough demonstrations per 
instruction to learn reliable behaviors and recovers the held-out combinations 
through compositional generalization.

\textbf{Sampling strategy shapes generalization behavior.} As shown in 
Fig.~\ref{fig:generalization_visualization_real_robot}, we visualize per-cell 
evaluation result on \textit{Bun} ($n=200$, $f=12$), where brown cells mark collected 
(training) combinations, and the number indicates the number of successful rollouts. \textsf{Random} scatters its budget across the grid, 
and its successes are correspondingly scattered. In contrast, \textsf{V} 
concentrates collection in a structured manner, yielding more successful 
generalization on the out-of-distribution cells.  This behavior is especially valuable in real-robot settings, where demonstration budgets are limited and each collected trajectory is costly; by avoiding exhaustive instruction coverage, bias-aware sampling has potential to scale to larger instruction spaces.


\begin{table*}[t]
\centering
\begin{minipage}[t]{0.68\textwidth}
\vspace{0pt}
\centering
\footnotesize
\setlength{\tabcolsep}{3.5pt}
\renewcommand{\arraystretch}{1.7}
\begin{tabular}{lcccc}
\toprule
Method 
& Bun(100)
& Bun(200)
& Pizza(100)
& Cup(100) \\
\midrule
\textsf{L} & 22.9\%(11/48) & 37.5\%(18/48) & 35.4\%(17/48) & 37.5\%(18/48) \\
\textsf{Random}        & 37.5\%(18/48) & 47.9\%(23/48) & 50.0\%(24/48) & 47.9\%(23/48) \\
\textsf{V(ours)}      & \textbf{54.2\%(26/48)} & \textbf{64.6\%(31/48)} & \textbf{62.5\%(30/48)} & \textbf{66.7\%(32/48)} \\
\bottomrule
\end{tabular}
\vspace{-6pt}
\caption{\textbf{Real-robot experiment results on GR00T-N1.7~\citep{bjorck2025gr00t}.}
The number in parentheses denotes the number of training demos.}
\label{tab:realrobot_experiment_result}
\end{minipage}
\vspace{-12pt}
\hfill
\begin{minipage}[t]{0.26\textwidth}
\vspace{0pt}
\centering
\includegraphics[width=\linewidth ]
{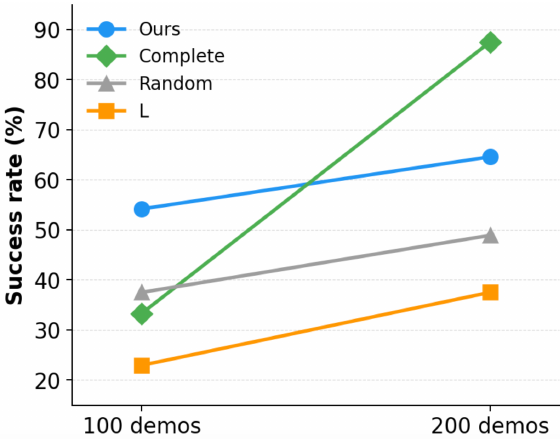}
\vspace{-15pt}

\captionof{figure}{Scaling behaviors on \textit{Bun}.}
\label{fig:real_world_scaling_behaviors}
\end{minipage}
\vspace{-5pt}
\end{table*}




\section{Conclusion}


In this work, we study instruction factor bias as a fine-grained shortcut failure in pretrained robot policies during finetuning. 
We show that policies often over-rely on dominant instruction factors while under-grounding others, limiting compositional generalization under out-of-distribution instructions. 
To quantify this behavior, we introduce factor-correlation experiments and propose Factor Dominance Rate (FDR) and Factor Dominance Hierarchy (FDH) for pairwise and global factor-bias diagnosis. 
Across six foundation policies, we find consistent patterns: color is the most dominant factor, while verb and size are often under-grounded. 
Motivated by this diagnosis, we propose a model-agnostic bias-aware data collection strategy that reallocates demonstrations toward under-grounded factors, showing that shaping data distributions to mitigate factor bias can improve compositional generalization beyond simply increasing data quantity and diversity.

\paragraph{Limitations and Future Works. } Our evaluation is conducted in tabletop manipulation settings with a controlled factor space; it remains an open question whether the identified factor bias patterns and our bias-aware strategy generalize to more complex open-world scenarios with richer object diversity or multi-step tasks. Additionally, our proposed \textsf{V} strategy is one instance of bias-aware allocation rather than a globally optimal strategy; future work could enhance our design through a principled search over coverage policies, e.g., via active learning or curriculum design. 




\clearpage






\appendix



 \renewcommand{\contentsname}{Table of Content}
  \setcounter{tocdepth}{3}   
  \tableofcontents
  \clearpage

\section{Factor Evaluation Space}

\label{app:instruction_space}

As described in Sec.3 (Page 3 in the main paper), every instruction is generated
from the template
\begin{quote}
\textit{``[Verb] the [Size] [Color] [Object] on the [Spatial Attribute] of the table.''}
\end{quote}
The full factor vocabulary is reproduced in Tab.~\ref{tab:factor_vocab}. The
Cartesian product of the five factors yields
$6\times6\times6\times6\times5 = 6{,}480$ unique instructions.

\begin{table}[h]
\centering
\small
\setlength{\tabcolsep}{6pt}
\caption{\textbf{Factor vocabulary.} Five instruction factors and their values.
We distinguish the \textit{action-centric} factor (Verb), which determines
\emph{how} the policy interacts with an object, from the four
\textit{object-centric} factors (Color, Object, Size, Spatial Attribute), which
determine \emph{which} object is selected and \emph{where}.}
\label{tab:factor_vocab}
\begin{tabular}{lll}
\toprule
\textbf{Factor} & \textbf{Type} & \textbf{Values} \\
\midrule
Verb               & action-centric & grasp, lift, push, pull, rotate, slide \\
Color              & object-centric & red, yellow, blue, orange, green, black \\
Object             & object-centric & cube, sphere, cup, car, pyramid, star \\
Size               & object-centric & small, large, smaller, larger, smallest, largest \\
Spatial Attribute  & object-centric & left, right, middle, front, behind \\
\bottomrule
\end{tabular}
\end{table}

\paragraph{Scene composition. } 
 Each scene instantiates the instructed target object together with a randomly sampled number of distractor objects, and is rebuilt at every reset so that no fixed object or spatial prior can be exploited. Implemented in ManiSkill on a Panda arm with a wrist camera (the resolution of agentic-view and inhand-view camera are both 256×256).

\section{Evaluation of Factor Bias and Factor Dominance Rate}
\label{app:factor_bias}

This section provides the full experimental details for the factor-bias
evaluation introduced in Sec. 3 of the main
paper. 


 



\paragraph{Training and Inference Instruction Details. }  As shown in Fig.~\ref{fig:factor_correlation_experiment_setups_visualization}(a), training language instructions are sampled across diagonal cells, where the evaluation language instructions are sampled across gray cells. More specifically, we list the instruction space of training and inference instruction as below. 

\paragraph{Training Instruction Space. }

\begin{itemize}
    \item \textsc{Color\,Object}: ``[Verb] the [color] [object].'' On-diagonal binds 6 (color, object) pairs: \emph{red+cube, yellow+sphere, blue+cup, orange+car, green+pyramid, black+star}. Carrier verb $\in\{\text{\emph{Grasp, Lift}}\}$. $\Rightarrow$ \textbf{12 training instructions}.
    \item \textsc{Color\,Spatial}: ``[Verb] the [color] cube at the [spatial].'' On-diagonal binds 5 (color, spatial) pairs: \emph{red+on the left, blue+in the middle, yellow+on the right, green+at the back, orange+in front}. Object fixed as \emph{cube}. Carrier verb $\in\{\text{\emph{Grasp, Lift}}\}$. $\Rightarrow$ \textbf{10 training instructions}.
    \item \textsc{Color\,Verb}: ``[Verb] the [color] [object].'' On-diagonal binds 6 (verb, color) pairs: \emph{lift+red, grasp+yellow, push+blue, pull+orange, rotate+green, slide+black}. Carrier object $\in\{\text{\emph{cube, sphere}}\}$. $\Rightarrow$ \textbf{12 training instructions}.
    \item \textsc{Color\,Size}: ``[Verb] the [size] [color] cube.'' On-diagonal binds 6 (color, size) pairs: \emph{red+small, blue+smaller, green+smallest, yellow+large, orange+larger, black+largest}. Object fixed as \emph{cube}. Carrier verb $\in\{\text{\emph{Grasp, Lift}}\}$. $\Rightarrow$ \textbf{12 training instructions}.
    \item \textsc{Object\,Spatial}: ``[Verb] the [object] at the [spatial].'' On-diagonal binds 5 (object, spatial) pairs: \emph{cube+on the left, cup+in the middle, sphere+on the right, pyramid+at the back, car+in front}. Carrier verb $\in\{\text{\emph{Grasp, Lift}}\}$. $\Rightarrow$ \textbf{10 training instructions}.
    \item \textsc{Object\,Verb}: ``[Verb] the [color] [object].'' On-diagonal binds 6 (verb, object) pairs: \emph{lift+cube, grasp+sphere, push+cup, pull+car, rotate+pyramid, slide+star}. Carrier color $\in\{\text{\emph{yellow, red}}\}$. $\Rightarrow$ \textbf{12 training instructions}.
    \item \textsc{Object\,Size}: ``[Verb] the [size] [object].'' On-diagonal binds 6 (object, size) pairs: \emph{cube+small, cup+smaller, sphere+large, pyramid+smallest, car+larger, star+largest}. Carrier verb $\in\{\text{\emph{Grasp, Lift}}\}$. $\Rightarrow$ \textbf{12 training instructions}.
    \item \textsc{Spatial\,Verb}: ``[Verb] the [object] at the [spatial].'' On-diagonal binds 5 (verb, spatial) pairs: \emph{lift+on the left, push+in the middle, grasp+on the right, rotate+at the back, pull+in front}. Carrier object $\in\{\text{\emph{cube, sphere}}\}$. $\Rightarrow$ \textbf{10 training instructions}.
    \item \textsc{Spatial\,Size}: ``Lift the [size] [object] at the [spatial].'' On-diagonal binds 5 (spatial, size) pairs: \emph{on the left+small, in the middle+smaller, on the right+large, at the back+smallest, in front+larger}. Verb fixed as \emph{Lift}. Carrier object $\in\{\text{\emph{cube, sphere}}\}$. $\Rightarrow$ \textbf{10 training instructions}.
    \item \textsc{Verb\,Size}: ``[Verb] the [size] [object].'' On-diagonal binds 6 (verb, size) pairs: \emph{lift+small, grasp+large, push+smaller, pull+larger, rotate+smallest, slide+largest}. Carrier object $\in\{\text{\emph{cube, sphere}}\}$. $\Rightarrow$ \textbf{12 training instructions}.
\end{itemize}

\paragraph{Evaluation Instruction Space. }
For Factor Evaluation Instruction Space, for each factor pair, only primary two factors are sampled from full grid, for example, in \textsc{Color, Object}, instructions are sampled from full grid which numerates color and object. Other values are sampled from  the same distribution.

\subsection{Experiment Setup.} For each of the pairwise factor correlation experiments, during training, 300 demonstrations are randomly sampled from the pairwise factor training instruction space. During evaluation, 400 rollouts are randomly performed under each evaluation instruction space. 

\subsection{Scoring with Gemini-2.5-Flash and Human Agreement}
\label{app:judge}

\paragraph{Why a VLM judge.} Rule-based success checks only report whether the
task was completed; they cannot attribute an \emph{overfitting failure} to a
specific factor. We therefore classify each agent-view rollout with
Gemini-2.5-Flash~\citep{comanici2025gemini} into one of:
\textit{success}, \textit{overfitting to $f_1$}, \textit{overfitting to $f_2$},
or \textit{other}. 

\paragraph{Why specifically choose Gemini-2.5-Flash. } Given the large number of rollout videos, for each models require 4000 videos to calculate FDR and FDH. Gemini-2.5-Flash
offers a favorable balance between annotation quality and throughput/cost.

\paragraph{Judge input and protocol. } For each rollout the judge receives the
instruction, a description of the two candidate objects in the scene, and the
recorded agent-view video; it returns the outcome label together with a short
justification.

\paragraph{Judge Model Prompt. } 

Our prompts are divided into \textbf{verb-conflict} template and \textbf{two-object} conflict template.

\textbf{Verb-conflict template}: Here are the four verb-primary experiments
(\textsc{Verb\,Color}, \textsc{Verb\,Object}, \textsc{Verb\,Size},
\textsc{Verb\,Spatial}):
\begin{verbatim}
Watch this robot manipulation video carefully.

**Instruction given to the robot:** "{instruction}"

This is a CONFLICT experiment about verb bias. The robot was trained
with specific verb-attribute pairings:
- Verb "{verb_i}" is paired with one attribute set (instruction verb)
- Verb "{verb_j}" is paired with {factor2_desc}

The robot must choose which action to perform. Observe carefully.

Answer in this EXACT format:
ROBOT_ACTION: [describe the physical action in 1 sentence]
ACTION_TYPE: ["{verb_i}" if instructed action, "{verb_j}" if competing
              action, "other" if neither]
FACTOR_FOLLOWED: ["verb" if robot followed instructed verb '{verb_i}',
                  "{factor2}" if robot followed the {factor2} factor
                  (doing '{verb_j}'), "neither" if unclear]
REASONING: [1-2 sentences explaining your judgment]
\end{verbatim}

\textbf{Two-object conflict template.} (the six non-verb-primary experiments
\textsc{Color\,Object}, \textsc{Color\,Size}, \textsc{Color\,Spatial},
\textsc{Object\,Size}, \textsc{Spatial\,Object}, \textsc{Spatial\,Size}):
\begin{verbatim}
Watch this robot manipulation video carefully.

**Instruction given to the robot:** "{instruction}"

This is an experiment about factor bias. The scene contains
two objects:
- Object A: has {factor1}='{val_i}'  -- matches '{factor1}' in instruction
- Object B: has {factor2}='{val_j}'  -- matches '{factor2}' in instruction

No object has BOTH attributes simultaneously. The robot must choose one.

Answer in this EXACT format:
ROBOT_ACTION: [describe what the robot arm did in 1 sentence]
OBJECT_TOUCHED: [A (matches {factor1}), B (matches {factor2}), neither]
FACTOR_FOLLOWED: ["{factor1}" if robot went for Object A,
                  "{factor2}" if Object B, "neither"]
REASONING: [1-2 sentences explaining your judgment]
\end{verbatim}




\paragraph{Output parsing. } Responses are field-parsed (regex on the five
\texttt{KEY: value} lines). Only \texttt{FACTOR\_FOLLOWED} feeds the FDR
statistic; the other fields support audit and qualitative inspection. Rollouts
labelled \texttt{neither} are excluded from $N_{f_1}+N_{f_2}$.

\noindent With $N_{f_1}$ and $N_{f_2}$ the counts of the two overfitting modes,
the Factor Dominance Rate is
\begin{equation}
    \text{FDR}(f_1, f_2) = \frac{N_{f_1} - N_{f_2}}{N_{f_1} + N_{f_2} + \epsilon}
    \in (-1, 1), \qquad \epsilon \rightarrow 0 .
\end{equation}

\paragraph{Agreement with human annotations. } To validate the judge, we compare
its labels against human annotations on a randomly sampled subset of rollouts. The judge achieves 88.7\% overall agreement with human annotations,
indicating strong consistency with human labels.

\begin{table}[h]
\centering
\small
\caption{\textbf{Agreement between the Gemini-2.5-Flash judge and human
annotations} on a held-out sample of rollouts of pi0.5.}
\label{tab:human_agreement}
\begin{tabular}{lcccc}
\toprule
& \textit{success} & \textit{overfit-$f_1$} & \textit{overfit-$f_2$} & \textbf{overall} \\
\midrule
Agreement (\%)        & 90.0 & 83.0 & 93.0 & 88.7  \\
\#\,annotated rollouts & \multicolumn{4}{c}{80} \\
\bottomrule
\end{tabular}
\end{table}


\subsection{Factor Dominance Hierarchy (FDH) via Copeland Ranking}
\label{app:fdh}

We aggregate the pairwise FDR scores into a global ranking via Copeland ranking.
Each factor $f_i$ receives a Copeland score:
\begin{equation}
    C(f_i) = \sum_{j \neq i} \mathbb{1}\!\left[\,\mathrm{FDR}(f_i, f_j) > \tau\,\right],
\end{equation}
where $\tau$ is a tie threshold (we use $\tau = 5\%$). A factor scores $+1$ for
each pairwise win and $0$ for each loss or tie; We report detailed experiment results of Copeland scores as indicated in Tab.~\ref{tab:copeland_tau5}.

\begin{table}[h]
\centering
\small
\setlength{\tabcolsep}{5pt}
\caption{Copeland scores $C(f_i)$ per model ($\tau=5\%$).}
\label{tab:copeland_tau5}
\begin{tabular}{lccccc}
\toprule
\textbf{Model} & Color & Object & Spatial & Verb & Size \\
\midrule
$\pi_{0.5}$         & 3 & 2 & 2 & 1 & 0 \\
$\pi_{0}$           & 4 & 2 & 1 & 1 & 0 \\
OpenVLA-oft         & 4 & 1 & 0 & 1 & 1 \\
GR00T-N1.7          & 4 & 3 & 2 & 1 & 0 \\
XVLA                & 3 & 3 & 1 & 1 & 0 \\
Genie-Envisioner    & 4 & 3 & 2 & 1 & 0 \\
\bottomrule
\end{tabular}
\end{table}

\begin{table}[h]
\centering
\small
\setlength{\tabcolsep}{5pt}
\caption{Copeland scores $C(f_i)$ per model ($\tau=10\%$).}
\label{tab:copeland_tau10}
\begin{tabular}{lccccc}
\toprule
\textbf{Model} & Color & Object & Spatial & Verb & Size \\
\midrule
$\pi_{0.5}$         & 3 & 1 & 0 & 1 & 0 \\
$\pi_{0}$           & 3 & 2 & 1 & 1 & 0 \\
OpenVLA-oft         & 3 & 1 & 0 & 1 & 1 \\
GR00T-N1.7          & 2 & 2 & 2 & 1 & 0 \\
XVLA                & 3 & 2 & 1 & 1 & 0 \\
Genie-Envisioner    & 4 & 3 & 2 & 1 & 0 \\
\bottomrule
\end{tabular}
\end{table}

\paragraph{Sensitivity of FDH to the tie threshold. } Moreover, we also report detailed experiment results of Copleland scores when as indicated in Tab.~\ref{tab:copeland_tau5} when $\tau = 5\%$ and Tab.~\ref{tab:copeland_tau10} $\tau = 10\%$. We also report the corresponding FDH as belows:

\textbf{At $\tau=5\%$:}
\begin{itemize}\itemsep0pt
    \item $\pi_{0.5}$: color $>$ \{object $\approx$ spatial\} $>$ verb $>$ size
    \item $\pi_{0}$: color $>$ object $>$ \{spatial $\approx$ verb\} $>$ size
    \item OpenVLA-oft: color $>$ \{object $\approx$ verb $\approx$ size\} $>$ spatial
    \item GR00T-N1.7: color $>$ object $>$ spatial $>$ verb $>$ size
    \item XVLA: \{color $\approx$ object\} $>$ \{spatial $\approx$ verb\} $>$ size
    \item Genie-Envisioner: color $>$ object $>$ spatial $>$ verb $>$ size
\end{itemize}

\textbf{At $\tau=10\%$:}
\begin{itemize}\itemsep0pt
    \item $\pi_{0.5}$: color $>$ \{object $\approx$ verb\} $>$ \{spatial $\approx$ size\}
    \item $\pi_{0}$: color $>$ object $>$ \{spatial $\approx$ verb\} $>$ size
    \item OpenVLA-oft: color $>$ \{object $\approx$ verb $\approx$ size\} $>$ spatial
    \item GR00T-N1.7: \{color $\approx$ object $\approx$ spatial\} $>$ verb $>$ size
    \item XVLA: color $>$ object $>$ \{spatial $\approx$ verb\} $>$ size
    \item Genie-Envisioner: color $>$ object $>$ spatial $>$ verb $>$ size
\end{itemize}

In summary, the findings across different models remain consistent, especially color $\geq$ object $\geq$ spatial $\geq$ verb $\geq$ size remains consistent across models and across different threshold. 

\section{Experiment}

\subsection{Different Data Collection Strategy}

We illustrate out data collection strategy as below. 

\begin{itemize}[leftmargin=*, topsep=2pt, itemsep=4pt]
  \item \textbf{\textsf{Complete}.} Collect from \emph{all} cells in the
  subspace, i.e., the full Cartesian product of factor values. This exhaustively
  covers the instruction space and serves as an upper-bound reference; it is
  infeasible at scale, especially on real robots. In our experiment, we report the performance of  \textbf{\textsf{Complete}} in real-robot experiment \textit{Bun} as the upper bound performance. 

  \item \textbf{\textsf{Random}.} Sample factor combinations uniformly at random
  over the full subspace until the budget $n$ is reached.

  \item \textbf{\textsf{L}.} Fix each factor at one value and sample the other
  factor(s) uniformly; for $N_f=2$ this collects one full row and one full
  column, forming an ``L'' shape on the grid.

  \item \textbf{\textsf{V(Ours).}} Following the intuition
  of~\citet{gao2024efficient}, we allocate coverage along the
  \emph{under-grounded} factor. For a pair $(f_1,f_2)$ with
  $\mathrm{FDR}(f_2,f_1) > \tau$ (i.e.\ $f_2$ dominates and $f_1$ is
  under-grounded), we collect the leftmost column---fixing $f_1$ and varying
  $f_2$---and additionally include the diagonal and subdiagonal of the grid to
  preserve joint coverage. For $N_f \geq 3$, one additional column is added per
  under-grounded factor, prioritizing the factors ranked lower by FDH.
\end{itemize}

We provide the pseudo code for \textbf{\textsf{L}} and \textbf{\textsf{V(Ours)
}} as below in the following tables when facing $N$ factors. 

\begin{algorithm}
\caption{\textsf{V} (Ours): bias-aware data collection for $N$ factors}
\label{alg:V}
\begin{algorithmic}[1]
\Statex \textbf{Input:} scene $\mathbf{S}$; factor grid $\mathbf{F}$ ($N$ factors $\times\, k$ values); base configuration $f^*$
\State $f \gets f^*$
\Statex \textit{// Phase 1 (N spokes): for each factor, sweep it alone with others at base}
\For{$i \gets 1$ \textbf{to} $N$}
    \State $f \gets f^*$ \Comment{reset between spokes}
    \For{$j \gets 1$ \textbf{to} $k$}
        \State $f[i] \gets \mathbf{F}_{i,j}$
        \State \textsc{SetFactors}$(\mathbf{S},f)$;\ \ \textsc{CollectData}$(\mathbf{S})$
    \EndFor
\EndFor
\Statex \textit{// Phase 2 (main diagonal): advance all factors together}
\For{$j \gets 1$ \textbf{to} $k$}
    \State $f[i] \gets \mathbf{F}_{i,j}$ \textbf{for all} $i \gets 1$ \textbf{to} $N$
    \State \textsc{SetFactors}$(\mathbf{S},f)$;\ \ \textsc{CollectData}$(\mathbf{S})$
\EndFor
\Statex \textit{// Phase 3 (N sub-diagonals): one factor one step ahead of the rest}
\For{$\ell \gets 1$ \textbf{to} $N$} \Comment{which factor leads}
    \For{$j \gets 1$ \textbf{to} $k-1$}
        \State $f[i] \gets \mathbf{F}_{i,j}$ \textbf{for all} $i \neq \ell$
        \State $f[\ell] \gets \mathbf{F}_{\ell, j+1}$
        \State \textsc{SetFactors}$(\mathbf{S},f)$;\ \ \textsc{CollectData}$(\mathbf{S})$
    \EndFor
\EndFor
\State \textbf{Keep unique configurations only}
\end{algorithmic}
\end{algorithm}

\begin{algorithm}
\caption{\textsf{L} Strategy Deployment}
\label{alg:L}
\begin{algorithmic}[1]
\Statex \textbf{Input:} scene $\mathbf{S}$, factor values $\mathbf{F}$ (size $N$ factors $\times\, k$ values), base factor values $f^*$ (size $N$ factors)
\For{$i \gets 1$ \textbf{to} $N$}
    \State $f \gets f^*$
    \For{$j \gets 1$ \textbf{to} $k$}
        \State $f_i \gets \mathbf{F}_{i,j}$
        \State \textsc{SetFactors}($\mathbf{S}, f$)
        \State \textsc{CollectData}($\mathbf{S}$)
    \EndFor
\EndFor
\State \textbf{Keep unique configurations only}
\end{algorithmic}
\end{algorithm}


\subsection{Factor-change counting per strategy}
\label{app:factor_change_per_strategy}
Let $N_f$ be the number of instruction factors in an experiment (e.g.\ $N_f{=}2$
for a factor pair, $N_f{=}5$ for the full space) and let $k_i$ be the number of
values of factor $i$ (here $k_i{=}6$ for verb, color, object and size, and
$k_i{=}5$ for the spatial factor). Following \citet{gao2024efficient}, the cost
of a collection schedule is the \emph{total number of factor changes}: each
transition between two consecutive configurations contributes its \textit{Hamming
distance} (the number of factors whose value changes; changing all $N_f$ factors
at once costs $N_f$).
\begin{itemize}[leftmargin=*, topsep=2pt, itemsep=2pt]
  \item \textbf{\textsf{Complete.}} visits all $\prod_{i=1}^{N_f} k_i$ factor
  combinations, costing $\mathcal{O}\!\big(\prod_i k_i\big)$ factor changes
  (e.g.\ $6{\times}6{=}36$ for a factor pair, and
  $6{\times}6{\times}6{\times}6{\times}5{=}6480$ for the full five-factor space)
  --- exponential in $N_f$ and infeasible at scale; used only as an upper-bound
  reference.
  \item \textbf{\textsf{V} (Ours).} starts from the base $f^*$ and
  varies one factor at a time in a staircase order, so each step changes exactly
  one factor and all values are covered in $\mathcal{O}\!\big(\sum_i k_i\big)$
  changes (linear). \textbf{Specifically}, when $N=2$, $N_f \approx 3*k$.
  \item \textbf{\textsf{L.}} starts from $f^*$ and lets each configuration deviate
  from $f^*$ by a single factor (a cross through $f^*$), also
  $\mathcal{O}\!\big(\sum_i k_i\big)$. \textbf{Specifically}, $N=2$, $N_f \approx 2*k$
\item \textbf{\textsf{Random.}} samples whole configurations IID; under independent uniform 
resampling of each factor, each transition has expected Hamming distance 
$\sum_{i=1}^{N_f} (k_i{-}1)/k_i$, so the expected total factor change after 
$n$ demonstrations is 
$\mathbb{E}[f_\text{Random}] = (n{-}1)\sum_i (k_i{-}1)/k_i = \mathcal{O}(nN_f)$, 
i.e.\ approximately $N_f$ changes per added demo.

\end{itemize}

Both composition-exploiting schedules (\textsf{V} and \textsf{L}) are linear,
$\mathcal{O}(\sum_i k_i)$, versus \textsf{Complete}'s exponential
$\prod_i k_i$. For fair comparison we fix the total factor change $f$ and the
budget $n$ across all methods.

Additionally, to match the total number of factor changes $f$ across strategies,
we sample a small number of additional demonstrations using the
\textbf{\textsf{Random}} strategy. Since this number is very small, it does not
substantially alter the original sampling distribution. We use this adjustment
because \textbf{\textsf{Random}} incurs a much higher factor-change cost,
especially in higher-dimensional settings.

\subsection{Simulation experiment settings}

\paragraph{Simulation platform. }
Simulation demonstrations are produced by scripted motion planning in
ManiSkill~\citep{mu2021maniskill} (\texttt{mplib}, per-environment solver), not
teleoperation or a learned policy. For each instruction configuration chosen by a
strategy, we build the factorized scene (target $+$ up to three randomized
distractors) on a Panda arm with a wrist camera and execute the instructed verb;
we retry until the target number of \emph{successful} demos is reached, discarding
failures. Each demo records an agent-view and a wrist image ($256\times256$) with
$8$-dim state and $8$-dim action, then is converted to LeRobot format. 

\paragraph{Simulation setup. } For each factor-pair experiment, the two primary factors vary on a $6\times6$
(or $6\times5$) on-diagonal grid; the carrier factor
appearing in the instruction template takes 2 randomly sampled values; the remaining
factors not in the instruction are fixed at default scene values. Scene-level
distractor objects (count and positions) are randomized per episode. Specific data sampling strategy is determined by $N_f$. By default, $N=200$ demonstrations are collected for training. During inference, by default 200 roll outs are performed, three random seeds are used for evaluation, we report each performance and their average.

\paragraph{Pretrained model checkpoints. }
We initialize all policies from their publicly released pretrained checkpoints
and post-train them on our collected demonstrations using each model's default
training configuration. All models are trained until convergence.

\subsection{Simulation experiment results}

\begin{wrapfigure}[11]{r}{0.4\linewidth}
    \vspace{-14pt}
    \centering

    \includegraphics[width=\linewidth]{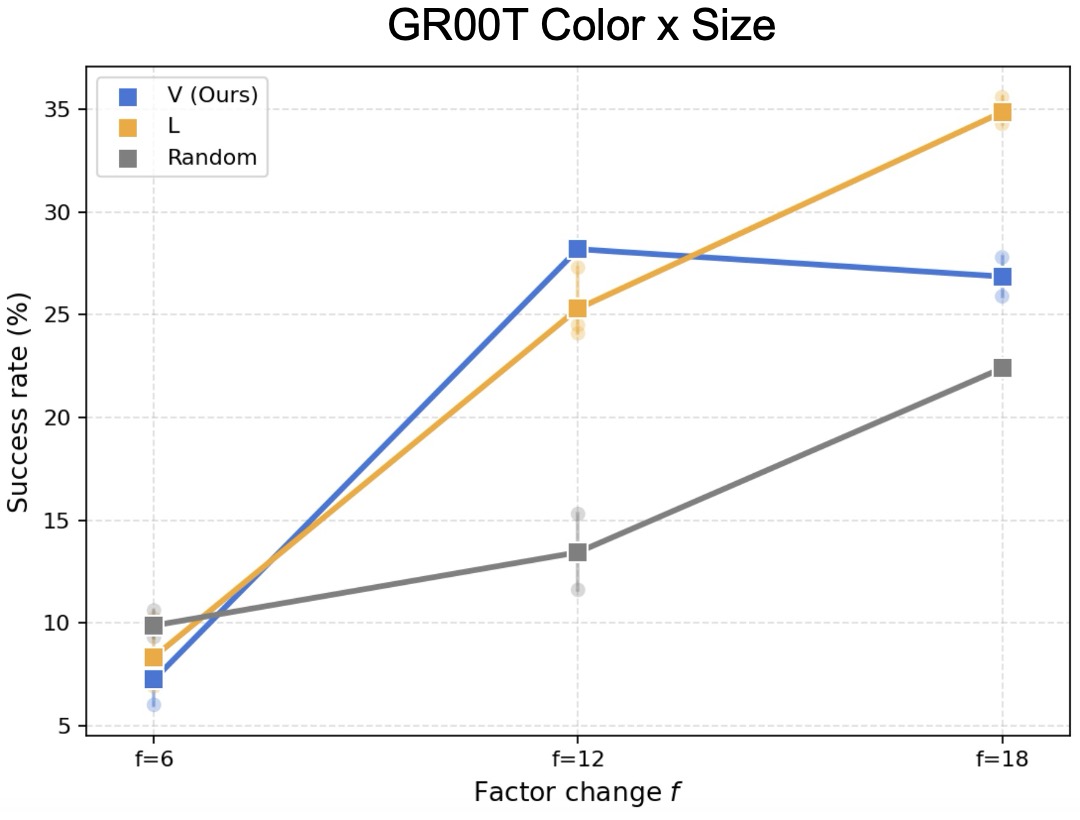}
    \vspace{-2pt}

    \caption{Additional Results.}
    \label{fig:data_experiment_result}
    \vspace{-10pt}
\end{wrapfigure}

We provide additional experiment results in Fig.~\ref{fig:data_experiment_result}, which are consistent with our main findings in two ways. First, bias-aware collection (V) brings the largest gains at \emph{higher} factor-change budgets: on \emph{Spatial$\times$Size} (GR00T) and \emph{Color$\times$Spatial} ($\pi_0$), V clearly surpasses both L and Random at $f{=}18$ ($48.1$ vs.\ $33.9/28.7$ and $54.9$ vs.\ $47.8/46.8$), whereas at small $f$ the three strategies are nearly indistinguishable. Second, on \emph{size}-involving pairs (\emph{Color$\times$Size}, \emph{Verb$\times$Size}) this advantage vanishes---L matches or exceeds V, and absolute success stays low---indicating that \emph{size} remains the hardest factor to ground irrespective of the collection strategy. This matches our factor-dominance analysis, in which size is the most under-grounded factor (lowest in the hierarchy): under a limited data budget no strategy reliably captures size, and only as the budget grows does structured allocation begin to help.

\subsection{Real-robot Experiment}

\paragraph{Tasks. } We evaluate the performance of our models on three real-robot tasks where multi-task language is naturally required. We provide details below, and additional experiment rollout videos are provided in supplementary materials. 

\textbf{\textit{Bun.}} In this task, the robot is asked to pick up a bun with specific color and place it into a specific location of the steamer.  Random distractors are placed near steamers and objects. The position of distractors are random. Instruction example: ``put (green, white, yellow, purple) bun into the (first, second, third, fourth) steamer basket on the left. ''

\textbf{\textit{Pizza.}} In this setting, robot is asked to put different sauce onto the Pizza. The positions of sauce are swapped with each oter. Instruction example: ``add (mustard sauce, toppings, mint, tomato sauce)'' to the pizza in the (top left, bottom right, top right, bottom left) 

\textbf{\textit{Cup.}} In this setting, robot is asked to put different sauce onto the Pizza. The positions of sauce are swapped with each other. Example language instrucion includes ``put the stirring spoon into the (blue, red, green, pink) cup'', ``serve the (pink, blue, red, green) cup on the tray'', ``shape the (pink, blue, red, green) cup'', ``pour out the water from (pink, blue, red, green) cup''

\paragraph{Evaluation Setup.} Each of the model is rolled out for 48 episodes. For each of the episode, the success condition is that the robot can accomplised the tasks in two trials.
Example template is ``Put the stirring spoon into the (pink, red, blue, green cup''
\clearpage

\bibliography{example}  

\end{document}